\documentclass{article}

\usepackage{nips_2017, float, graphicx}

\usepackage[utf8]{inputenc} 
\usepackage[T1]{fontenc}    
\usepackage{hyperref}       
\usepackage{url}            
\usepackage{booktabs}       
\usepackage{amsfonts}       
\usepackage{nicefrac}       
\usepackage{microtype}      
\usepackage{caption}
\usepackage{sidecap}

\title{Topology and Prediction Focused Research on\\ Graph Convolutional Neural Networks}

\author{
  Matthew Baron\\
  Electrical & Computer Engineering Department\\
  Carnegie Mellon University\\
  Pittsburgh, PA 15213 \\
  \texttt{mcbaron@cmu.edu} \\
}

\begin{document}

\maketitle

\begin{abstract}
Important advances have been made using convolutional neural network (CNN) approaches to solve complicated problems in areas that rely on grid structured data such as image processing and object classification. Recently, research on graph convolutional neural networks (GCNN) has increased dramatically as researchers try to replicate the success of CNN for graph structured data. Unfortunately, traditional CNN methods are not readily transferable to GCNN, given the irregularity and geometric complexity of graphs. The emerging field of GCNN is further complicated by research papers that differ greatly in their scope, detail, and level of academic sophistication needed by the reader.  

The present paper provides a review of some basic properties of GCNN. As a guide to the interested reader, recent examples of GCNN research are then grouped according to techniques that attempt to uncover the underlying topology of the graph model and those that seek to generalize traditional CNN methods on graph data to improve prediction of class membership. Discrete Signal Processing on Graphs (DSP$_G$) is used as a theoretical framework to better understand some of the performance gains and limitations of these recent GCNN approaches.  A brief discussion of Topology Adaptive Graph Convolutional Networks (TAGCN) is presented as an approach motivated by DSP$_G$ and future research directions using this approach are briefly discussed.   

\end{abstract}

\section{Introduction}
A wide variety of scientific and engineering fields use graphs or network structures to represent complex data that describe interactions between objects of interest. Transportation routes between cities, relationships in a social network, or biomarkers for neurological disorders are just a few examples of data that can be structured in graph form. Graphs are particularly useful for describing irregular or non-Euclidean geometric structures of data domains. A graph is characterized by a set of nodes or vertices and a set of edges between pairs of vertices. The weight associated with each edge in a graph typically represents the similarity between the two vertices it connects. 

Convolutional Neural Network (CNN) approaches routinely achieve better than human accuracy in tackling complex problems in areas which involve image processing and object classification wherein large scale datasets are represented in grid form [1]. However, the irregularity and geometric complexity of graph data pose many challenges for traditional CNN approaches. The potential of CNN approaches to solve complicated problems using graph data has sparked an explosion of recent papers on graph convolutional neural networks (GCNN). The rate of new research on GCNN is measured in months rather than years. 

In the present paper, I describe some basic properties of GCNN and review some recent approaches in GCNN research. I categorize recent GCNN techniques based on whether the goal of the approach is to 1) better describe the underlying model that produced the graph data or if the primary goal is to 2) better directly predict class membership for new data. Some of the performance gains of these GCNN approaches are then highlighted using the theory of Discrete Signal Processing on Graphs (DSP$_G$) [2]. Concepts related to Topology Adaptive Graph Convolutional Networks (TAGCN), an approach largely motivated by the theory of DSP$_G$, are then described [3]. Future directions for research using TAGCN are discussed.

\section{Graph Structured Representations of the Real World}
\subsection{Notation}
First, let us define some notation used throughout this paper. A graph is represented $G = (\mathcal{V}, \mathcal{E})$ where $\mathcal{V}$ is the vertex set with $|\mathcal{V}| = n$ and $\mathcal{E}$ is the set of all edges. We denote $\mathbf{A}$ as the adjacency matrix and $\mathbf{D}$ the degree matrix of $\mathbf{A}$ $ \mathbf{D}_{ii} = \sum_j \mathbf{A}_{ij}$. The graph Laplacian matrix is defined as $\mathbf{L} = \mathbf{D} - \mathbf{A}$, the normalized Laplacian, $\hat{\mathbf{L}} = \mathbf{D}^{-\frac{1}{2}} \mathbf{A} \mathbf{D}^{-\frac{1}{2}}$. The Laplacian is diagonalized by the Fourier basis $ \mathbf{U} = [ \ldots ] \in \mathbb{R}^{n \times n}$ so that $\mathbf{L} = \mathbf{U} \Lambda \mathbf{U}^\top$ For the sake of consistency, layers will be indexed by $t$ and the output of the $t$th layer notated as $H^{(t)}$. 

\subsection{Properties}
Network data that represent real world phenomena typically have properties that allow us to make some simplifying analytical assumptions [4]. Real world networks are generally representable with a sparse graph in which the number of edges is not quadratic in the number of nodes, but linear. The sparseness of real world graphs also manifests degree heterogeneity, meaning that a few vertices are highly connected while most vertices have few connections. Most vertices belong to a single component in which there exists a path between any two nodes. Additionally, many real networks lack bipartite components which implies that the maximum eigenvalue of the symmetric Laplacian matrix is 2 [5]. This property is leveraged in existing literature as an approximation for computational simplification [6, 7, 8]. In real world graphs, weights on the edges are generally real, non-negative, and frequently are directed, meaning the adjacency matrix is not symmetric. Additionally, graphs in the real world exhibit stationarity, meaning the statistics of the stochastic process do not vary in time, or that any trends are consistent and can be removed. 

\subsection{Graph Convolutional Neural Network (GCNN) Approaches}
Recently, there has been a rapid proliferation of papers on different types of GCNN techniques. For example, 53\% of arXiv research papers that reference Kipf and Welling’s seminal work on graph convolutional networks [6] have been written since January 2018. Although the rate of new research is testimony to the potential of GCNN, recent papers in this area differ greatly in their scope, level of detail, and assumptions about the academic background of readers. One goal of the present paper is to offer a way to group different research approaches to GCNN for further comparison and to highlight examples of these different techniques for the interested reader.  

The overarching goal of GCNN models is to predict or classify nodes in a graph using a subset of labeled nodes and node features. Researchers are ultimately interested in developing GCNN approaches that can achieve fast network training times and high predictive accuracy. Recent research on GCNN can be grouped into work that primarily focuses on GCNN techniques for obtaining more information about the structure of the manifold being sampled and research that uses GCNN techniques to more effectively leverage information in the observed graph features for prediction or classification. The work of Segarra and colleagues [9] on topology inference, an Attention-based Graph Neural Network technique proposed by Thekumparampil et al. [10] and Kipf and Welling’s Variational Graph Auto-Encoders [11] are all approaches that use information from observed graph features to augment and refine an understanding the underlying graph topology. In contrast, models that introduce new forms of neural network operations such as the graph convolution and pooling proposed by Defferrard et al. [7] and influence propagation techniques described by Kipf and Welling [6] focus on the prediction or classification of unlabeled nodes on a graph. 

\section{GCNN to Model Underlying Graph Topology}
To model the underlying topology of graph data, a first topology focused category of GCNN techniques consider ways to predict the existence of edges in the graph. By concentrating on the observed data, these methods expose the underlying manifold structure that can enable a better understanding of the information propagation dynamics of the network. Let us briefly consider three different approaches to obtaining information about the underlying topology of the manifold based on observed data. 

Segarra et al. [9] use concepts from graph signal processing (GSP) to make inferences about the relational structure of the network by leveraging the observed graph signals. Their statistical method for graph topology inference views the observed graph signals as the outputs of an unknown graph filter that is intended to be learned. Segarra et al. take the observed feature values on the nodes of the graph as the response of a linear network diffusion process to a variety of inputs. With sufficiently many observed feature vectors and some assumed constraints on the network as a system, the approach solves for the adjacency matrix (i.e., the topology of the system). Specifically, the approach assumes that the graph signals resulted from the diffusion dynamics of the network and that the signals are graph stationary, and that eigenvectors of the graph shift operator can be estimated from the sample covariance. \begin{equation}
\mathbb{E}[xx^\top] = \mathbf{U} diag(|\hat{h}|^2) \mathbf{U}^\top
\end{equation}
The adjacency matrix can be determined from an eigendecomposition of the graph shift operator. 

Research by Thekumparampil et al. [10] describes neural network-based methods to predict links from node features and to infer some weighting of the connections. Their Attention-based Graph Neural Networks (AGNN) approach is effectively augmenting a binary adjacency matrix with information about the relative influence of neighboring nodes, given information about the graph signal. The output row-vector of node $i$ at layer $t+1$ is given by 
\begin{equation}
H_i^{(t+1)} = \sum_{j \in N(i) \cup \{i\}} P_{ij}^{(t)} H_{j}^{(t)}
\end{equation}
 The propagation matrix P at each layer of the network is an adaptively learned weighting scheme for the neighborhoods of each node based on the cosine distance between the responses of the previous layer to the features at that node. \begin{equation}
P_{ij}^{(t)} = \frac{1}{C} e^{\beta^{(t)} cos(H_i^{(t)}, H_j^{(t)})}
\end{equation}
This captures how relevant node $j$ is to node $i$ at the scale represented at layer $t$. AGNN constructs a non-binary adjacency matrix which it uses as a graph shift operator with a learned gain term. 

Kipf and Welling’s Variational Graph Auto-Encoders (VGAE) [11] framework takes a somewhat different approach to modeling the underlying structure of the graph network. Rather than attempting to find connections not explicitly made in observed graph data, VGAE uses the concept of the Variational Auto-Encoder to find a latent variable space which can completely represent an undirected graph. VGAE does not rely on an adjacency matrix as given. The VGAE approach learns a representational transformation via the collection of observed graph signals that can reconstruct an adjacency matrix $\hat{\mathbf{A}}$ via a simple inner product. $\hat{\mathbf{A}} = \sigma(\mathbf{Z} \mathbf{Z}^\top)$ Using a learned, reconstructed adjacency matrix means that the model is more robust to errors and higher order noise in the raw adjacency matrix. Since many phenomena in the real world can be represented in a low dimensional space, the VGAE approach of finding the latent variable space and using it as a noise reduction mechanism makes it very attractive for achieving error reduction. 

Research approaches that focus on topology inference are useful in model refinement which can lead to improved predictive performance. As we have seen, Segarra et al. [9] and Thekumparampil et al. [10] take approaches to discover the topology of the manifold by learning the graph shift matrix directly or by augmenting a binary graph shift matrix with edge weights. The encoding method of VGAE does not learn or augment a graph shift matrix, although one can be reassembled from the embedded latent space. These three methods are useful as a data-explorational processing stage and the learned graph shift can be used in conjunction with the prediction focused category of GCNN techniques to increase accuracy of classification or prediction. 

\section{GCNN to Predict Class Membership}
A second category of GCNN approaches focuses on directly improving the predictive capabilities of neural networks for graph data structures. These models integrate the connectivity patterns and feature attributes of graph structured data in a novel manner for prediction using strategies analogous to those found in traditional CNN. Defferrard [7] discusses the steps required to generalize CNNs in the design of convolutional filters on graphs. Kipf and Welling [6] propose a simple information propagation rule for graph data derived from a first-order approximation of spectral graph convolution.

Defferrard et al. [7] introduce a novel approach to the pooling operation on graphs. Pooling is an important operation for images, or any data that exhibits stationarity, as it allows features  that are influential in one region to be influential to other regions as well. These graph coarsening and pooling operations rely on an assumption of the graph signals being locally stationary and compositional. As the authors show, this assumption is true for certain types of graph constructions such as those on a Euclidean grid (i.e. images) and less true for others, such as a bag of words model for text categorization. The authors do not define graph convolution in the vertex domain, citing an inability to “express a meaningful translation operation” (p. 3). Instead, the convolution operator is defined in the Fourier domain such that $x \ast_{G} y = U((U^\top x) \odot (U^\top y))$ and a signal $x$ filtered by $g_\theta$ is
\begin{equation}
g_\theta(\mathbf{L})x = g_\theta(U\Lambda U^\top) x = U g_\theta(\Lambda) U^\top x
\end{equation} In addition, the authors define spectral filtering on graphs in this spectral (Fourier) domain, as well as a polynomial parameterization for localized filters. To lower the computational complexity of their formulation, the recursive Chebyshev polynomial approximation is introduced to avoid multiplication by the Fourier basis of eigenvectors. $g_\theta(\Lambda) \approx \sum_k \theta_k T_k(\Lambda)$ This Chebyshev approximation is leveraged in a number of GNN models proposed by other researchers, including Kipf and Welling [6]. 

One of the most influential papers in this prediction focused category of GCNN approaches is Kipf and Welling's “Semi-Supervised Classification with Graph Convolutional Networks.” The authors describe a fast, layer-wise linear approximation of the spectral graph convolution defined in Defferrard et al. [7]. Beginning with the Chebyshev polynomial approximation to spectral convolution \begin{equation}
g_\theta \ast x \approx \sum_{k = 0}^{K} \theta_k T_k(\tilde{\mathbf{L}}) x
\end{equation}
 Kipf and Welling restrict $K = 1$, looking only at filters which are a linear function on the graph Laplacian. This constraint may help prevent over-fitting, but limits the size and representational power of the learned filters. Lastly, the authors ensure no bipartite components in the graph by adding the identity to the symmetrically normalized graph Laplacian, effectively adding an extra self-loop to each node. This step is key to the validity of the Chebyshev polynomial expansion, making 2 an upper-bound on the maximum eigenvalue of the graph Laplacian matrix. 

As Kipf and Welling [6] note, their Graph Convolutional Network (GCN) approach suffers from a reduction in accuracy with increased depth of the graph convolution layers. The authors propose a solution in the appendix of the paper that uses residual or skip connections which allow information to pass from earlier layers to deeper layers unaltered. I posit that the features from these skip connections are given more importance in the deeper layers than those from the preceding layer because the power of the GCN lies not in the nonlinearity, but in the design of the label propagation mechanism. The importance of the skip connections for information propagation is further validated by the work of Zhao et al. [8] in their adaptation of common image identification convolutional network topologies to include graph convolutions. These researchers compare the topologies of a 6-layer baseline, the DenseNet framework which concatenates each layer of convolved features, and the ResNet framework with skipped connections. The authors maintained the neural network topology for each framework, switching out image convolutions for the Chebyshev approximated graph filters of GCN to generate a G(raph)\_ResNet and G(raph)\_DenseNet. Results suggest that the G\_ResNet framework out performed both the G\_DenseNet and baseline topologies for the tested social network datasets. These findings suggest that to improve accuracy of deep GCNN structures, it is necessary to include features calculated from shallower layers. This phenomenon might be a consequence of the expressions used for graph convolution, as discussed in the next section. 

An example of the effect the model quality of the neural network topology is available in Li et al. [12]. The authors show the substantial effect that adding information propagation in the form of graph convolution has over a simple fully connected network. Without a mechanism for mixing the information present in node features, the decrease in accuracy on an example citation network dataset is on the order of 20\%. It is evident that an implementation of Laplacian smoothing is needed to achieve high-accuracy graph neural network topologies.

\section{Discrete Signal Processing on Graphs: Theoretical Insights}
We have seen that GCNN research can be grouped according to techniques that attempt to uncover the underlying topology of the graph model and approaches that seek to generalize traditional CNN methods on graph data to improve prediction of class membership. Techniques that focus on the underlying graph topology can be used to advance the predictive capacities of GCNN at the data-explorational processing stage. We have described approaches that focus on approximations to the operation of graph convolutional such as those proposed by Defferrard et al. [7] and by Kipf and Welling [6]. Let us now briefly consider a theoretical framework proposed by Sandryhaila and Moura [2] to examine some of the advantages and limitations of these filter models in GCNN research.

In their work on Discrete Signal Processing on Graphs (DSP$_G$), Sandryhaila \& Moura generalize classical discrete signal processing concepts to extent to graph structured signals. The authors begin their paper by discussing the classical DSP filter structure of a tapped delay line in a more general algebraic context. Sandryhaila \& Moura note that filters are given as polynomials in the shifting operation which in the case of graph signals is multiplication by the adjacency matrix. Filtering using a tapped delay line is a time-domain operation; therefore, the equivalent algebraic structure for graph filtering is a vertex domain operation and not a spectral one. In this formulation, a graph filter $\mathbf{H}$ is a polynomial in $\hat{\mathbf{A}}$
\begin{equation}
\mathbf{H} = h(\hat{\mathbf{A}}) = h_0 \mathbf{I} + h_1 \hat{\mathbf{A}} + \dots + h_L\hat{\mathbf{A}}^L
\end{equation}
where $\hat{\mathbf{A}}$ is the normalized adjacency matrix to ensure that $\mathbf{H}$ is computationally stable, ie the eigenvalues of $\hat{\mathbf{A}}$ and the poles of $\mathbf{H}$ are within the unit circle. 

Using this DSP theoretical framework, we can examine some of the expressions that arise in the literature about graph convolutional filters. Let us return to Kipf and Welling’s [6] first order construction of the GCN in which the expression of the propagation model is $X \Theta_0 + D^{-\frac{1}{2}} A D^{-\frac{1}{2}} X \Theta_1$. Careful examination reveals that this expression is simply a linear polynomial in the chosen graph shift matrix. Let us also consider their renormalization expression. $\tilde{D}^{-\frac{1}{2}} \tilde{A} \tilde{D}^{-\frac{1}{2}} X \Theta$ Kipf and Welling cite their renormalization trick $I_N + D^{-\frac{1}{2}} A D^{-\frac{1}{2}} \rightarrow \tilde{D}^{-\frac{1}{2}} \tilde{A} \tilde{D}^{-\frac{1}{2}} $ with $\tilde{A} = A + I_N$ and $ \tilde{D}_{ii} = \sum_j \tilde{A}_{ij}$ as a simplification meant to alleviate the projection along the graph eigenvector of largest eigenvalue. The authors demonstrate a 1-2\% improvement over the first order spectral formulation of the GCN when evaluated against the Citeseer, Cora and Pubmed datasets. However, this improvement is a consequence of the form of the network. Effectively, Kipf and Welling have re-parameterized the entire GCN as a network to be an adaptive cascade of linear filter sections as defined in Sandryhaila \& Moura [2] \begin{equation}
h(\mathbf{A}) = (\mathbf{I}_N + h_t \mathbf{A}) (\mathbf{I}_N + h_{t-1} \mathbf{A}) \dots (\mathbf{I}_N + h_1 \mathbf{A})
\end{equation} 

The argument against non-linearities is made indirectly in Thekumparampil et al. [10] as they explore the basis for their attention mechanism. By removing the element-wise non-linearity which Kipf and Welling require as part of their construction of a GCN, Thekumparampil et al. [10] define a Graph Linear Network (GLN) as 
\begin{equation}
Z = f(X, A) = softmax((P^2X) W^{(0)} W^{(1)})
\end{equation} With $P = \tilde{D}^{-\frac{1}{2}} \tilde{A} \tilde{D}^{-\frac{1}{2}}$ as in GCN.
Examining the form of this expression more closely, we can see that in training the GLN learns a weight matrix for a two-hop network, as the matrix expression simplifies to $softmax(P^2 X W') $. Repeated left multiplication by the chosen graph shift operator simply expands the application of the shared weights W to not only adjacent nodes, but also the 2-neighborhood of each node. In effect this expression allows for a quadratic term in the filter polynomial, but no linear term. Thekumparampil et al. [10] report that this GLN matches the performance $\pm$ .3\% of GCN on the CiteSeer, Cora and PubMed citation network datasets. 

A number of the formulations for graph convolution proposed in the approaches just described rely on symmetric and regularized graph Laplacian matrices as the graph shift operator. This parameterization choice does not respect that real-world networks are frequently directed, nor does it respect graphs with negative edge weights. As Sandryhaila and Moura [2] state, “In general, the graph Laplacian is a second-order operator for signals on a graph, whereas an adjacency matrix is a first-order operator. Deriving a graph Fourier transform from the graph Laplacian is analogous in traditional DSP to restricting signals to be even (like correlation sequences) and Fourier transforms to represent power spectral densities of signals (p. 2).” The networks in the common reference datasets represent real, directed, and possibly asymmetric graphs. As a result, relying on the graph Laplacian matrix in the definition of the graph convolution is a restriction of the representational power of the models that researchers have trained. The theory from classical DSP supports using the adjacency matrix in the construction of graph convolution for the most generalizable filter constructions. However, depending on what is desirable in application specific contexts, the geometry of the underlying manifold may be better characterized by a discrete approximation to another operator (i.e. a Schrödinger-type) beyond the Laplace-Beltrami. 

Researchers that propose techniques that exploit the propagational features of graph structured data for the sake of improved predictive accuracy note that over-mixing or over-smoothing is the fundamental issue that arises from repeated multiplication of a graph shift matrix [12]. In the case of a deep GCN or a deep GLN, the output of the last graph convolutional layer is simply the projection of the output of the first convolutional layer along the eigenvector corresponding to the largest magnitude eigenvalue of the graph shift matrix. This is true because the expression for a graph filter reduces to a scalar weighted monomial of the shift matrix in both GCN and GLN. In other words, the input graph features are over-smoothed, since the deep network is performing power iteration as an eigenvector algorithm. If the expression of the graph filter is not solely a monomial in the graph shift matrix, and if the degree of the polynomial is varied in successive layers, then the problem of over-smoothing is avoided. 

\section{Topology Adaptive Graph Convolutional Networks}
We have discussed ways that the theoretical framework of DSP$_G$ can provide insights into current approaches to predicting class membership or feature values using GCNN.  In this final section of this paper, we take a brief look at Topology Adaptive Graph Convolutional Networks (TAGCN), a GCNN technique that draws directly from DSP$_G$ concepts.  Future research directions using a TAGCN approach are discussed.

In their paper, Topology Adaptive Graph Convolutional Networks, Du et al. [3] propose a generalization of the GCN design to use different size filters in each layer of the convolutional network. According to the theory outlined in DSP$_G$, the vertex domain graph filter formulation allows for an arbitrary size (or length) polynomial in the graph shift operator. This reframing avoids the linearized approximation of convolution via Chebyshev polynomial expansion in the Kipf and Welling GCN framework, thus enhancing representational capability. Although there are double the number of weights to be learned per layer, TAGCN does not require a computationally intensive eigendecomposition of the graph Laplacian matrix as proposed by Defferrard [7]. By generalizing the construction of the graph convolution, the approach becomes consistent with classical signal processing and can fit into other CNN architectures. 

A future direction for research on TAGCN is to examine the effect of implementing dropout on the predictive capabilities of the approach. The principle of dropout was first discussed by Srivastava et al. [13] for neural networks and involves randomly disconnecting learned weights during training. The idea behind implementing dropout in TAGCN is to take advantage of the varied topologies of the neural network that arise. The varied topology training allows parts of multiple neural networks to work together to reduce error, also known as boosting or model averaging. 

Examining the expression used in learning the weights of each graph filter in TAGCN offers an appropriate place in which to implement dropout. \begin{equation}
H^{(t+1)} = ReLU(\mathbf{A}^K H^{(t)} W_K^{(t)} + \dots + \mathbf{A}^1 H^{(t)} W_1^{(t)} + H^{(t)} W_0^{(t)} + \mathbf{B})
\end{equation} Dropout on the matrix W implements model averaging across the coefficients of the graph filter polynomial. 
Another area in the construction of TAGCN to allow dropout is directly on the higher powers of the adjacency matrix A. Although A is a sparse matrix due to the nature of the real-world network it represents, successive powers of A become increasing dense as connections are made to larger neighborhoods. Dropping out connections to larger neighborhoods can reduce the effect of the over-mixing caused by repeated multiplication by the graph shift matrix. This operation can be thought of in the classical DSP sense as filtering with a filter that has time varying coefficients.

\section{Conclusions}
In this paper, we presented a way to categorize recent research on Graph Convolutional Neural Networks as being either topology focused, or prediction focused.  The aim of topology focused methods is to learn more about the manifold that supports the observed collection of signals, either by directly approximating and augmenting the adjacency matrix, or via a learned latent space representation.  Prediction focused methods introduce and justify mathematical expressions that generalize the convolution operator to graph structured signals.  Advances and limitations of these prediction focused methods were discussed in the context of the theory of Discrete Signal Processing on Graphs.  The recent Topology Adaptive Graph Convolutional Network approach was briefly described as a technique based on Discrete Signal Processing on Graphs.

This paper was meant to offer some insights into navigating recent research in GCNN for the interested reader and was in no way intended to be a comprehensive review of the literature.  Examples of topology focused and prediction focused methods were selected to be illustrative but other research examples could just have easily been chosen to reflect current trends in the literature.  Moreover, discussion of important concepts in machine learning were ignored such as distinctions between unsupervised, semi-supervised, and supervised learning. The paper also did not address the advantages and limitations of using a combination of topology focused and prediction focused methods in specific applications.  Clearly, GCNN approaches hold much potential for helping to solve complex scientific, engineering, and even social problems in meaningful ways.  It is an exciting time for research in this emerging field.

\section*{References}
\small
[1] Y. LeCun, Y. Bengio, and G. Hinton. Deep learning. \textit{Nature}, 521: 436 – 444.\\

[2] A. Sandryhaila and J. M. F. Moura. Discrete signal processing on graphs. \textit{IEEE Transactions on Signal Processing}, 31(5), 2013.\\

[3] J. Du, S. Zhang, G. Wu, J M. F. Moura, and S. Kar. Topology adaptive graph convolutional networks. \textit{arXiv preprint arXiv: 1710.10370}, 2018.\\

[4] P. Latouche and F. Rossi. Graphs in machine learning: An introduction. \textit{arXiv preprint arXiv: 1506.06962}, 2015.\\

[5] F. R. K. Chung and F. C. Graham. \textit{Spectral Graph Theory}. American Mathematical Society, 1997.\\

[6] T. N. Kipf and M. Welling. Semi-supervised classification with graph convolutional networks. In \textit{Proceedings of the International Conference on Learning Representations (ICLR)}, 2017.\\

[7] M. Defferrand, X. Bresson, and P. Vandergheynst. Convolutional neural networks on graphs with fast localized spectral filtering. \textit{Proceedings of the Conference on Neural Information Processing Systems (NIPS)}, 2016.\\

[8] W. Zhao, C Xu, Z. Cui, T. Zhang, J. Jiang, Z. Zhang, and J. Yang. When work matters: Transforming classical network structures to Graph CNN. \textit{arXiv preprint arXiv: 1807.02653}, 2018.\\

[9] S. Segarra, A. G. Marques, G. Mateos, and A. Ribeiro. Network topology inference from spectral templates. \textit{IEEE Transactions on Signal and Information Processing over Networks}, 3(3):467 – 483.\\

[10] K. K. Thekumparampil, C. Wang, S. Oh, and L. Li. Attention-based graph neural network for semi-supervised learning. \textit{arXiv preprint arXiv: 1803.03735}, 2018.\\

[11] T. N Kipf and M. Welling. Variational graph auto-encoders. \textit{arXiv preprint arXiv: 1611.07308I}, 2016.\\

[12] Q. Li, Z Han, and X. Wu. Deeper insights into graph convolutional networks for semi-supervised learning. \\

[13] N. Srivastava, G. Hinton, A. Krizheysky, I Sutskever, and R. Salakutdinov. Dropout: a simple way to prevent neural networks from overfitting. \textit{Journal of Machine Learning Research}, 15: 1929-1958, 2014.\\

\end{document}